\documentclass[10pt, conference, compsacconf]{IEEEtran}
\IEEEoverridecommandlockouts
\usepackage{cite}
\usepackage{amsmath,amssymb,amsfonts}
\usepackage{algorithm}
\usepackage{algpseudocode}
\usepackage[pdftex]{graphicx}
\usepackage{textcomp}
\usepackage{here}
\newcommand{\argmax}{\operatornamewithlimits{argmax}}

\def\BibTeX{{\rm B\kern-.05em{\sc i\kern-.025em b}\kern-.08em
    T\kern-.1667em\lower.7ex\hbox{E}\kern-.125emX}}
\begin{document}

\title{Faster Deep Q-learning \\ using Neural Episodic Control
}

\author{
\IEEEauthorblockN{Daichi Nishio}
\IEEEauthorblockA{\textit{Institute of Science and Engineering} \\
\textit{Kanazawa University}\\
\textit{Kanazawa, Japan}\\
\textit{Email: dnishio@csl.ec.t.kanazawa-u.ac.jp}}
\and
\IEEEauthorblockN{Satoshi Yamane}
\IEEEauthorblockA{\textit{Institute of Science and Engineering} \\
\textit{Kanazawa University}\\
\textit{Kanazawa, Japan}\\
\textit{Email: syamane@is.t.kanazawa-u.ac.jp}}
}
\maketitle

\begin{abstract}
The research on deep reinforcement learning which estimates Q-value by deep learning has been attracted the interest of researchers recently.
In deep reinforcement learning, it is important to efficiently learn the experiences that an agent has collected by exploring environment.
We propose NEC2DQN that improves learning speed of a poor sample efficiency algorithm such as DQN by using good one such as NEC at the beginning of learning.
We show it is able to learn faster than Double DQN or N-step DQN in the experiments of Pong.

\end{abstract}


\section{Introduction}

Deep Q-Network (DQN)\cite{DQN} have made a success of deep reinforcement learning end-to-end, and various algorithms have been proposed since then\cite{survey}.
However learning a task with large state space is difficult, and many learning steps are necessary especially in an environment where rewards are sparse.

\begin{table}[htbp]
\caption{The differences between DQN and NEC}
\begin{center}
\begin{tabular}{l|c|c}
\hline
\textbf{ } & \textbf{DQN} &\textbf{NEC}\\
\hline
Estimation & Neural Network & Neural Network + DND\\
\hline
Sampling & Random or with Priority & Random\\
\hline
Memory cost & Large & Huge\\
\hline
\end{tabular}
\label{tab0}
\end{center}
\end{table}

In order to solve them, it is necessary to efficiently use experiences obtained by exploration.
DQN uses Experience Replay\cite{ER} which stores experiences in memory called Replay Buffer and It trains with minibatch randomly.
Prioritized Experience Replay\cite{PriER} has been proposed to learn more efficiently than random sampling.
It considers experiences with large train error as important experiences, and greatly improves learning speed and performance.

Neural Episodic Control (NEC)\cite{NEC} is another way to efficiently learn.
It uses the memory module called Differentiable Neural Dictionary (DND) to learn stably with a smaller number of learning steps.
The agent can decide its action by taking advantage of past similar experiences stored in DND.
In addition, it is able to learn end-to-end because it is differentiable inside a neural network.

However, DND needs a large memory for each action.
It also requires a lot of calculation time and memory usage.
We show their relations in Table \ref{tab0}.

In this research, we propose a method of improving learning efficiency and speed by adapting NEC's learning efficiency to a simple network like DQN.

\section{Deep Reinforcement Learning}


We target reinforcement learning assuming general Markov Decision Process(MDP).
We define the state of the environment at time $t$ as $s_t$, and the agent selects the action $ a_t $ by the policy $\pi$.
Then it obtains the reward $r_t$ and the next state $s_{t+1}$ corresponding to $ a_t $ from the environment.
The revenue is discounted return $G_t = \sum_t (\gamma^t r_t)$, where $\gamma$ is discount rate as the degree of consideration of the future.
The action-value function the agent uses for selecting action is defined as $Q_\pi (s, a) = E_\pi [G_t \mid s, a] $.
The optimal Q-value is based on the Bellman optimal equation \cite{bellman} as follows.
\begin{equation}
Q^*(s, a) = E[ r + \gamma \max_{a'} Q(s', a')\mid s,a ]\label{Q*}
\end{equation}

Q-learning\cite{Q-learn} is used to obtain the optimal Q-value.
\begin{equation}
Q(s, a) \gets Q(s, a) + \alpha( r + \gamma \max_{a'} Q(s', a') - Q(s, a))\label{Q-learning}
\end{equation}

In Q-learning, assuming that samples $(s, a, r, s')$ can be obtained infinitely from all pairs of $(s, a)$, \eqref{Q-learning} obtains the optimal Q-value function $ Q^*(s, a) $. 
Also, it converges to the same value since the Q-value function does not depend on the policy.
On the other hand, the convergence may take a long time if there are pairs $(s, a)$ that are not tried.

In DQN, the agent uses the $\varepsilon$-greedy policy for tradeoff between exploration and exploitation.
This $\varepsilon (0 < \varepsilon < 1)$ is a constant, or there is a method of linear decaying as increasing learning steps.
\begin{equation}
\pi (a\mid s) = \left \{
\begin{array}{l}
1-\varepsilon \ \ (a = \argmax_a Q(s, a)) \\
\ \ \ \varepsilon \ \ \ \ (otherwise)
\end{array}
\right.
\end{equation}

DQN aiming at learning from images uses Convolutional Neural Network (CNN)\cite{CNN} as a state feature extractor. 
The Q-value for embedding $h$ obtained from CNN is estimated by fully-connected layers.

The agent stores tuples of $(s_t, a_t, r_t, s_{t+1})$ in Replay Buffer so that it can learn with a minibatch formed randomly from the buffer.
In addition, for stability of the target value, it uses a target network for calculating a target value separately from a neural network for learning.
The target network uses parameter $\theta^-$ which is slightly older than the current parameter $\theta$ of the learning network.

A neural network learns from the loss function $L(\theta) = E[y_t - Q(s, a;\theta)]$ using the target value in \eqref{target}.
\begin{equation}
y_t = r_t + \gamma \max_{a'} Q(s_{t+1}, a';\theta^-)\label{target}
\end{equation}

Hasselt et al.\cite{Double Q-learn} show that DQN overestimates the action-value when the number of experience samples obtained from a environment is small, and Double Q-learning is a solution to it.
Double Q-learning updates the Q-value using two Q-value estimators A and B.
\begin{equation}
Q^A(s, a) \gets Q^A(s, a) + \alpha (r + \gamma Q^B(s', a^*) - Q^A(s, a))\label{QA}
\end{equation}
\begin{equation}
Q^B(s, a) \gets Q^B(s, a) + \alpha (r + \gamma Q^A(s', b^*) - Q^B(s, a))\label{QB}
\end{equation}
\[ where \ a^*= \argmax_a Q^A(s', a), b^* = \argmax_a Q^B(s', a) \]

$Q^A $ prevents $Q^B$ from overestimated, and $Q^B$ prevents $Q^A$ from it, respectively.

The algorithm called Double DQN\cite{DDQN} using Double Q-learning for DQN has been also proposed.
Equation \eqref{target} is rewritten as follows.
\begin{equation}
y_t = r_t + \gamma Q(s_{t+1}, \argmax_a Q(s_{t+1}, a;\theta);\theta)\label{target_}
\end{equation}

Hasselt et al. have incorporated the idea of Double Q-learning into \eqref{target_}.
\begin{equation}
y_t^{Double} = r_t + \gamma Q(s_{t+1}, \argmax_a Q(s_{t+1}, a;\theta);\theta^-)\label{target_DDQN}
\end{equation}

This makes it possible to get higher scores with 90\% of games played by DQN, and it is still widely used as an better algorithm than DQN.

\section{Related work}

Neural Episodic Control(NEC) is one of the method of efficient sampling from Replay Buffer.
It is the algorithm based on episodic memory and improved Model-Free Episodic Control (MFEC)\cite{MFEC} to learn end-to-end from state mappings to estimations of Q-values.

Differentiable Neural Dictionary(DND) has been proposed to make it successful.
DND for a action $a \in A$ is a dictionary $M_a = (K_a, V_a)$ which saves a pair of key $K_a$ and value $V_a$.
The key is the embedding $h$ which is the feature extracted the state $s \in S$ with CNN, and the value is the Q-value.

We can perform two kinds of operations for DND, $Lookup$ and $Write$.
In $Lookup$, when $ h $ featured by CNN and corresponding action $ a $ are entered,
we lookup the top $p$-nearest neighbors for $ h $ in $M_a$ using kd-trees \cite{kd}.
We weight the value $ v_i $ corresponding to $p$ keys and we set it as $ Q_a $. 
\begin{equation}
w_i = \frac{k(h, h_i)}{\sum_j k(h, h_j)}\label{wi}
\end{equation}
\begin{equation}
Q_a = \sum_i w_i v_i\label{Qa}
\end{equation}

$k(h, h_i)$ is a kernel function for $h$ and $h_i$.
In DND, \eqref{kernel} is used.
\begin{equation}
k(h, h_i) = \frac{1}{\|h-h_i\|_2^2+\delta}\label{kernel}
\end{equation}

Although $ \delta $ is a parameter to prevent division by zero, we should make it little larger such as $\delta = 10^{-3} $ because each value of $p$-nearest neighbors is referred to a certain extent.

In $Write$, we write an input $ h $ and a corresponding Q-value in DND.
If the key that already matches the input $ h $ exists in $ M_a $, update the corresponding Q-value according to the following with a learning rate $\alpha$.
\begin{equation}
Q_i \gets Q_i + \alpha (Q^{(N)}(s, a) - Q_i)\label{Qi}
\end{equation}

When the size of the dictionary reaches the upper limit, we overwrites a pair that has not been referred to recently as the top $p$-nearest neighbor value according to Least Recently Used (LRU).

NEC uses N-step Q-learning\cite{multi-step} as a target value.
\begin{equation}
Q^{(N)}(s_t, a) = \sum_{j=0}^{N-1} \gamma^jr_{t+j} + \gamma^N \max_{a'} Q(s_{t+N}, a')\label{NstepQ}
\end{equation}

However, N-step Q-learning is hard to be stabilized by off-policy algorithm\cite{N-step Safe}.

We use NEC in our proposed algorithm, but we replace the output Q-value of NEC network with the Q-value defined in Chapter \ref{PA}.

\section{Proposed Algorithm} \label{PA}

\begin{figure*}[t]
\centering
\includegraphics[width=14cm]{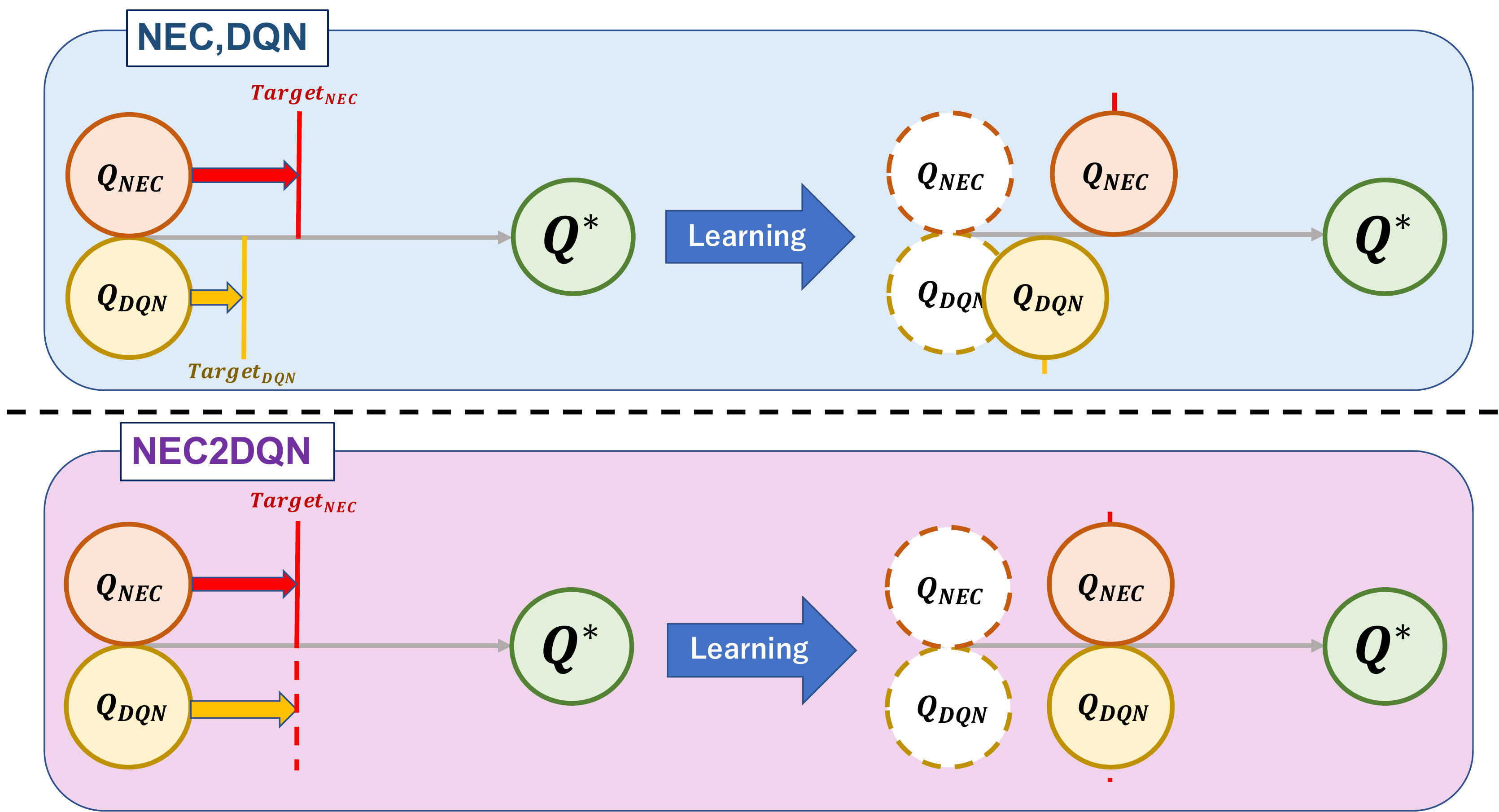}
\caption{
The image of the convergence of NEC2DQN: $Q_{NEC}$ and $Q_{DQN}$ will converge to the same optimal state-action value $Q^*$. 
The top shows the progress of convergence when NEC and DQN learn respectively. 
The bottom is the progress of convergence of each Q-value by using NEC2DQN.
Although $ Q^* $ is unknown and cannot be observed directly, NEC and DQN should be close to $Q^*$.
In the early stage of learning, it is better to use the $Target_{NEC}$ as shown in the figure.
Therefore, we want DQN to learn by approaching not $Target_{DQN}$ but $Target_{NEC}$.
}
\label{nearQ}
\end{figure*}

NEC currently has the following problems.

\begin{enumerate}

\item \label{enu} As the number of pairs of key and value in DND increase, the computation time for finding top $p$-nearest neighbors increases.\\

\item \label{eenu} The number of dictionaries $M$ of DND is the same as the size of the action space $ |A| $.\\

\item \label{eeenu} State space and action space need to be discrete because Q-learning is used.\\

\end{enumerate}

Regarding \ref{eeenu}, Matsumori et al. \cite{Keio} are addressing research in an environment where the state space is continuous and Partially Observable Markov Decision Process (POMDP).

Our work focuses on the problems \ref{enu} and \ref{eenu}. 
Continuing to use NEC requires many computational resources due to constraints of time computational quantity and space computational quantity.
Therefore, we will address using NEC's sampling efficiency only in early learning of other deep Q-learning algorithms.
We will use DQN as an example which is the simplest Deep Q-learning network and we call this algorithm NEC2DQN(N2D).

As mentioned in \cite{bellman}, there is always one optimal action-value $ Q^* $, and if it is the same policy, it always converges to the same value.
For this reason, both DQN algorithm and NEC algorithm are possible to head to the same $ Q^* $. 
Therefore, the Q-value estimated by other algorithms can be taken as the target value.
Hence it is easier to converge by using a better target value, the value approaches to $ Q^* $ faster. 
We show this simple image in Figure \ref{nearQ}.

We set $Q^A(s, a)=Q_{DQN}(s, a)$, $Q^B(s, a)=Q_{NEC}(s, a)$ in \eqref{QA} and \eqref{QB}. 
Then we rewrite them as follow.
\begin{equation}
\small Q_{DQN}(s, a) \gets Q_{DQN}(s, a) + \alpha (r + \gamma Q_{NEC}(s', a^*) - Q_{DQN}(s, a))\label{QA'}
\end{equation}
\begin{equation}
\small Q_{NEC}(s, a) \gets Q_{NEC}(s, a) + \alpha (r + \gamma Q_{DQN}(s', b^*) - Q_{NEC}(s, a))\label{QB'}
\end{equation}
\[where \  a^*= \argmax_a Q_{DQN}(s', a) , b^* = \argmax_a Q_{NEC}(s', a) \]

\begin{figure*}[t]
\centering
\includegraphics[width=13cm]{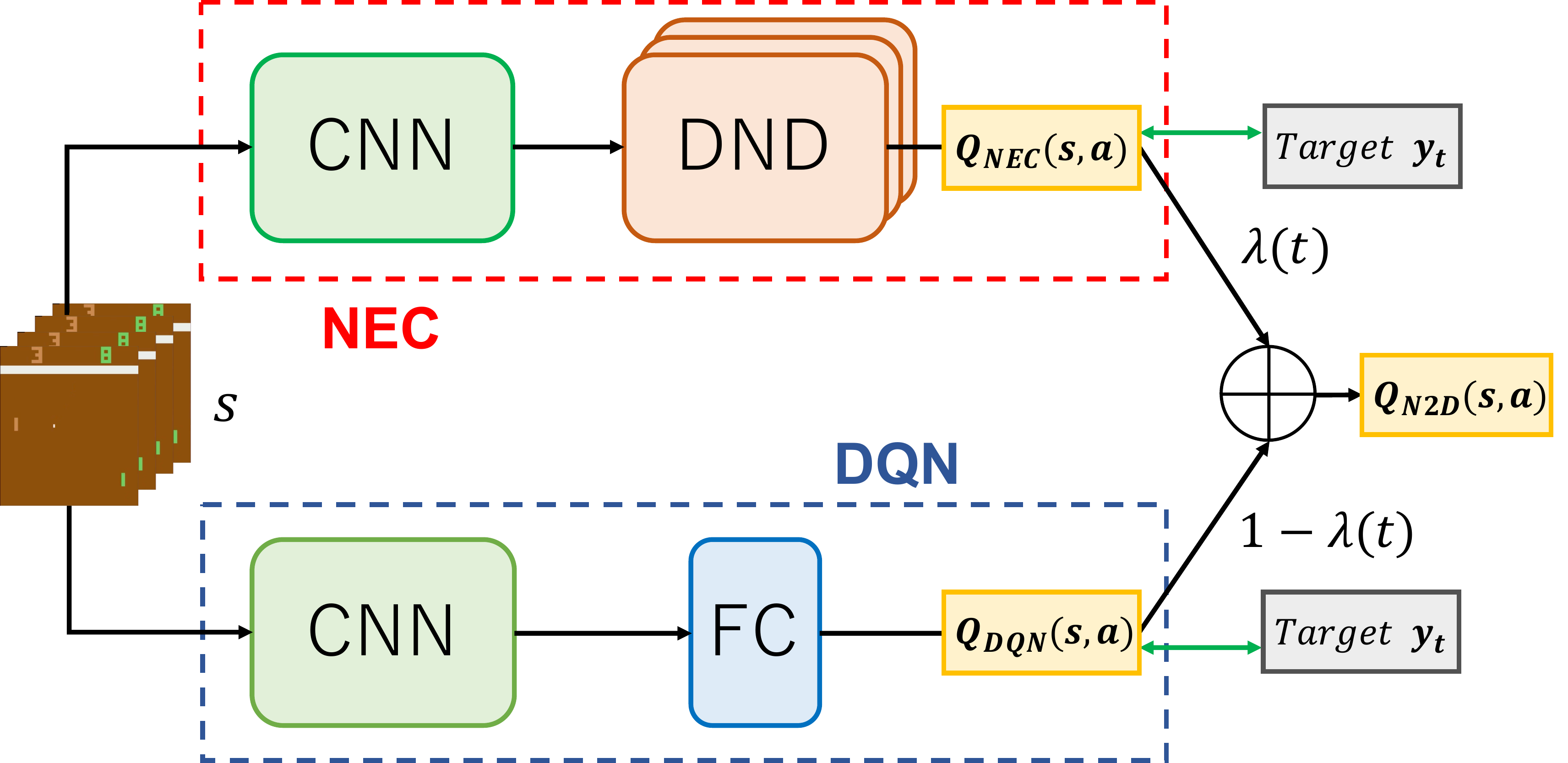}
\caption{NEC2DQN Network}
\label{Network}
\end{figure*}

Double DQN is one-step Q-learning, but NEC uses N-step returns. Thus we want to rewrite \eqref{QA'} and \eqref{QB'} like N-step Double DQN.
Instead of preparing a target network, NEC stores the target value $ Q_{NEC}^{(N)} $ in Replay Buffer.
In order to fit this target value, we adopt N-step returns for DQN(hereinafter, we call it N-step DQN).
\begin{equation}
Q_{DQN}(s, a) \gets Q_{DQN}(s, a) + \alpha(Q_{NEC}^{(N)}(s, a^*) - Q_{DQN}(s, a))\label{QA''}
\end{equation}
\begin{equation}
Q_{NEC}(s, a) \gets Q_{NEC}(s, a) + \alpha(Q_{DQN}^{(N)}(s, b^*) - Q_{NEC}(s, a))\label{QB''}
\end{equation}

Equations \eqref{QA''} and \eqref{QB''} show $Q_{NEC}^{(N)}(s, a^*)$ is necessary for learning $Q_{DQN}(s, a)$, and $Q_{DQN}^{(N)}(s, b^*)$ is necessary for  learning $Q_{NEC}(s, a)$.
However, $ Q_{NEC} $ may not be able to learn well until DQN learns as a network, and there is a possibility that $ Q_{DQN} $ cannot be learned well due to the influence.
Especially, NEC should be able to learn earlier by learning using the original $ Q_{NEC}^{(N)} $.
Therefore we rewrite \eqref{QB''} like \eqref{QB'''}.
\begin{equation}
Q_{DQN}(s, a) \gets Q_{DQN}(s, a) + \alpha(Q_{NEC}^{(N)}(s, a^*) - Q_{DQN}(s, a))\label{QA'''}
\end{equation}
\begin{equation}
Q_{NEC}(s, a) \gets Q_{NEC}(s, a) + \alpha(Q_{NEC}^{(N)}(s, b^*) - Q_{NEC}(s, a))\label{QB'''}
\end{equation}

$ Q_ {DQN} (s, a) $ can learn with a better target value by NEC algorithm at the beginning of learning.
Even if learning this way, $ Q_ {DQN} $ converges to the same value as $ Q_ {NEC} $.
But we need to consider the learning has advanced because this research aims to use NEC only in the early stage of learning.
Since $Q_{DQN}$ maybe estimated correctly to some extent, it can be learned like Double DQN using N-step returns.
\begin{equation}
Q_{DQN}(s, a) \gets Q_{DQN}(s, a) + \alpha(Q_{DQN}^{(N)}(s, a^*) - Q_{DQN}(s, a))\label{QA''''}
\end{equation}
\begin{equation}
Q_{NEC}(s, a) \gets Q_{NEC}(s, a) + \alpha(Q_{DQN}^{(N)}(s, b^*) - Q_{NEC}(s, a))\label{QB''''}
\end{equation}

Comparing \eqref{QA'''} with \eqref{QB'''}, they use the same target value $Q_{NEC}^{(N)}$.
Similarly, both \eqref{QA''''} and \eqref{QB''''} use $Q_{DQN}^{(N)}$.
Hence we replace them with the same Q-value $Q_{N2D}^{(N)}$.
\begin{equation}
Q_{DQN}(s, a) \gets Q_{DQN}(s, a) + \alpha(Q_{N2D}^{(N)}(s, a^*) - Q_{DQN}(s, a))\label{QA'''''}
\end{equation}
\begin{equation}
Q_{NEC}(s, a) \gets Q_{NEC}(s, a) + \alpha(Q_{N2D}^{(N)}(s, b^*) - Q_{NEC}(s, a))\label{QB'''''}
\end{equation}

From here, we define $ Q_{N2D} (s, a) $ to satisfy the above property.
We prepare networks of NEC and DQN separately as shown in Figure \ref{Network}.
Both of the networks output their Q-value for the state $s_t$, and we combine them as $ Q_{N2D}(s_t, a) $. 
\begin{equation}
Q_{N2D}(s_t, a) = \lambda(t) Q_{NEC}(s_t, a) + (1-\lambda(t))Q_{DQN}(s_t, a) \label{Q_N2D}
\end{equation}

This $ \lambda (t) $ is a monotonously decreasing function related to the current learning step $t$. 
It represents how much to consider NEC.
We make it linear decay from 1 to 0 as increasing learning steps such as \eqref{lambda}.
\begin{equation}
\lambda(t) = \left \{
\begin{array}{l}
1-\frac{t}{CS} \ \ (t < CS) \\
\ \ \ 0 \ \ \ \ \ \ \  (otherwise)
\end{array}\label{lambda}
\right.
\end{equation}

At the beginning of learning, it refers to the Q-value of NEC, and gradually refers to the Q-value of DQN as learning progresses.
Thus it is possible to switch naturally.

We set the steps $CS$ to start to fully depend on DQN.
It is not necessary to calculate $ Q_ {NEC} (s_t, a) $, and the calculation time is also reduced because $ \lambda(t) = 0 $ after $CS$.

Loss functions of NEC and DQN are required respectfully because they are separate networks, but we use the same target value $ y_t $.
Although we use $a^*$ and $b^*$ in \eqref{QA'''''} and \eqref{QB'''''}, we do not use Double Q-learning but simply use N-step returns since we do not want to use $Q_{N2D}^{(N)}(s, a^*)$.
\begin{equation}
y_t = Q_{N2D}^{(N)}(s, a) = \sum_{j=0}^{N-1} \gamma^j r_{t+j} + \gamma^N \max_{a'} Q_{N2D}(s_{t+N}, a')\label{Target_N2D}
 \end{equation}

This is based on the fact that it is hard for overestimation of the Q-value to occur when the amount of experience accumulated in Replay Buffer is small because NEC can refer to DND.

These loss functions also use same target $y_t$. 
\begin{equation}
L_{NEC}(\theta_t) = E[y_t - Q_{NEC}(s_t, a_t)]
\end{equation}
\begin{equation}
L_{DQN}(\theta_t) = E[y_t - Q_{DQN}(s_t, a_t)]
\end{equation}

Similarly to original NEC, Replay Buffer stores a tuple of $ (s_t, a_t, y_t) $.
We accumulate the trajectory of the episode at the end of each episode because $ y_t $ requires N steps of reward data and the subsequent state $ s_{t + N} $.

We show the overall algorithm in Algorithm \ref{alg1}.

\begin{algorithm}[htbp]
\caption{NEC2DQN}
\label{alg1}
\begin{algorithmic}[1]
\State Initialize the number of entire timesteps $S$ to 0.
\State Initialize the change step $CS$ for $\lambda(t)$
\State Initialize replay memory $D$ to capacity $C_D$.
\State Initialize DND memories $M_a$ to capacity $C_{M_a}$.
\State Initialize action-value function $Q_{NEC}$ and $Q_{DQN}$ with random weights.
\For{each episode}
\State Initialize trajectory memory $G$ .
\For{$t = 1,2,...;T$} \Comment{Explore and train.}
\State Receive observation $s_t$ from environment.
\State Receive $Q_{DQN}(s_t, a)$.
\If{$S < CS$} \Comment{If $S < CS$, we use NEC.}
\State Receive embedding $h$ and $Q_{NEC}(s_t, a)$.
\Else
\State Set $Q_{NEC}(s_t, a)$ to free values.
\EndIf
\State Calculate $Q_{N2D}(s_t, a)$. \Comment{Calculate by \eqref{Q_N2D}.}
\State $a_t \gets \varepsilon$-greedy policy based on $Q_{N2D}(s_t, a)$.
\State Take action $a_t$, receive reward $r_t$.
\State Append $(s_t, a_t, r_t)$ to $G$.
\State Train on a random minibatch from $D$.
\State $S \gets S + 1$
\EndFor
\For{$t = 1,2,...;T$} \Comment{Calculate N-step returns.}
\State Calculate $y_t$. \Comment{Calculate by  \eqref{Target_N2D}.}
\State Append $(s_t, a_t, y_t)$ to $D$.
\If{$S < CS$}
\State Append $(h_t, y_t)$ to $M_{a_t}$.
\EndIf
\EndFor
\EndFor
\end{algorithmic}
\end{algorithm}

\begin{figure*}[t]
\begin{minipage}{0.5\hsize}
\centering
\includegraphics[width=5.0cm]{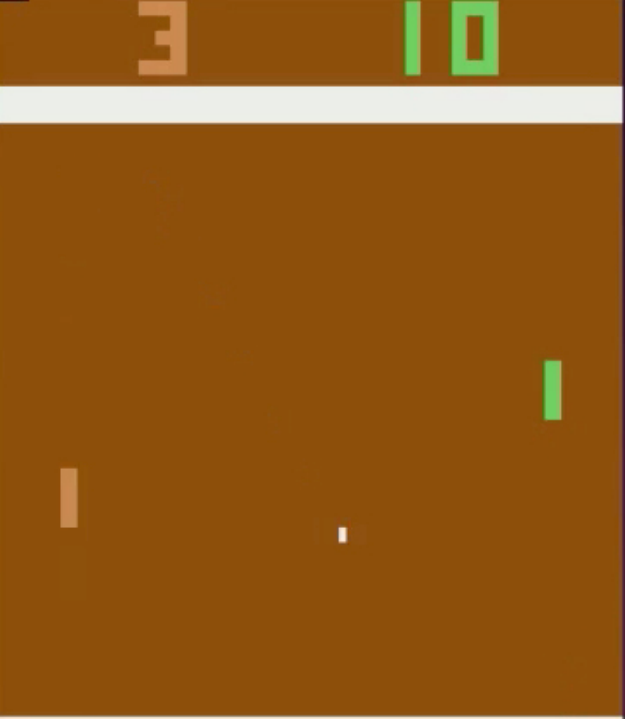}
\caption{A Pong frame}
\label{Pong}
\end{minipage}
 \begin{minipage}{0.5\hsize}
\centering
\includegraphics[width=9cm]{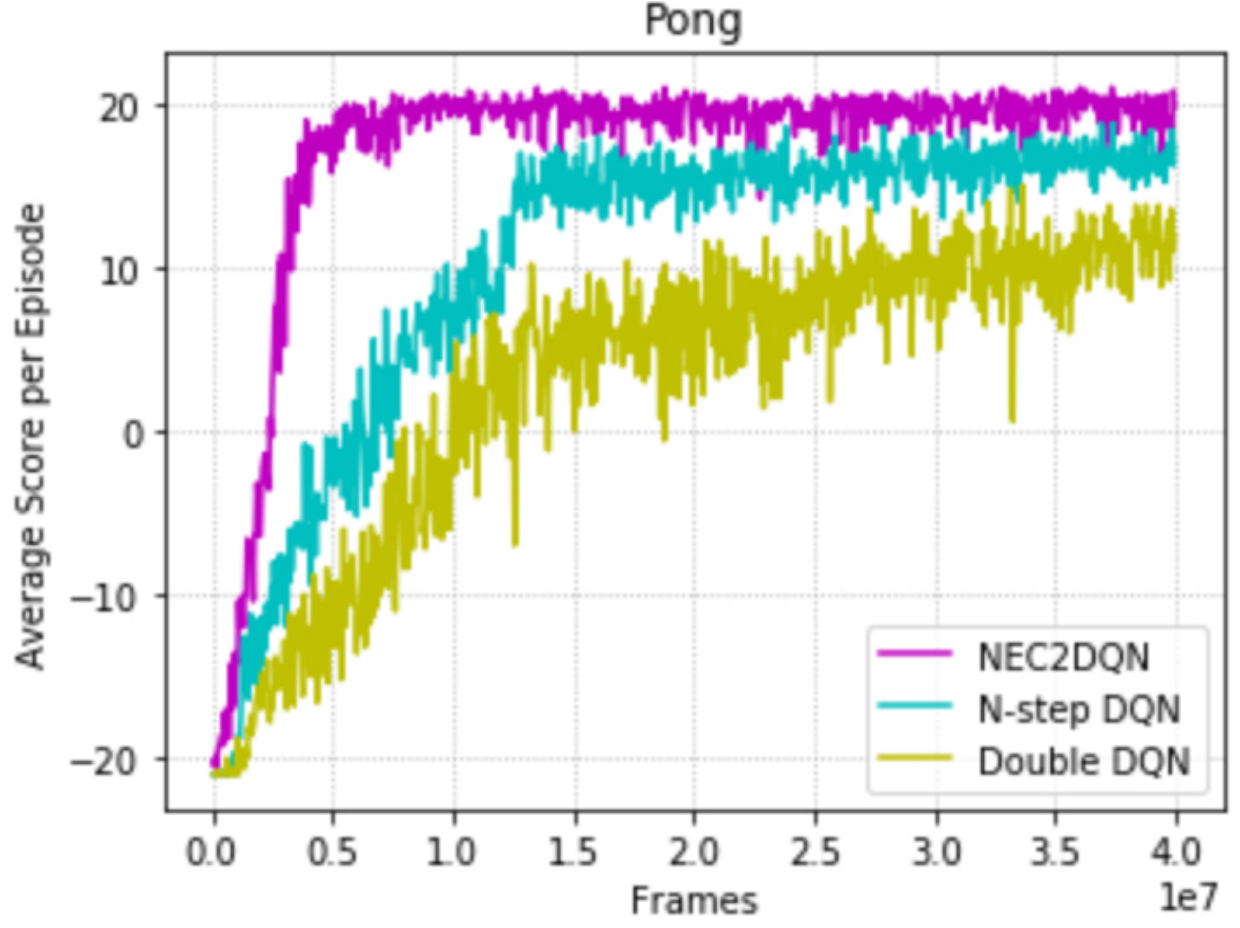}
\caption{The comparing NEC2DQN with other algorithms}
\label{Others}
 \end{minipage}
 
 \begin{minipage}{0.5\hsize}
  \centering
  \includegraphics[width=9cm]{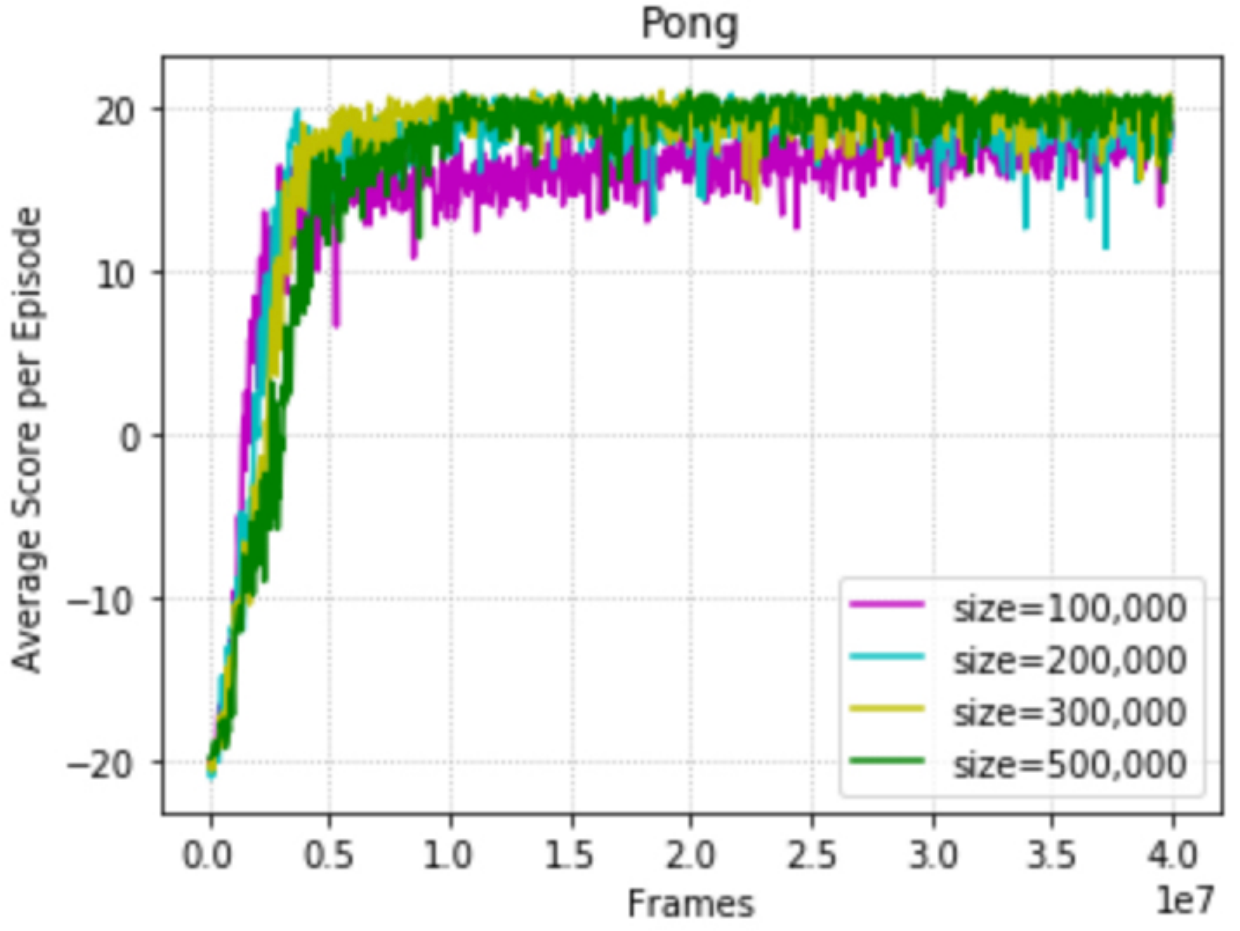}
  \caption{The difference of NEC2DQN performance by Replay Buffer size}
  \label{Buffer} 
 \end{minipage}
 \begin{minipage}{0.5\hsize}
 \centering
  \includegraphics[width=9cm]{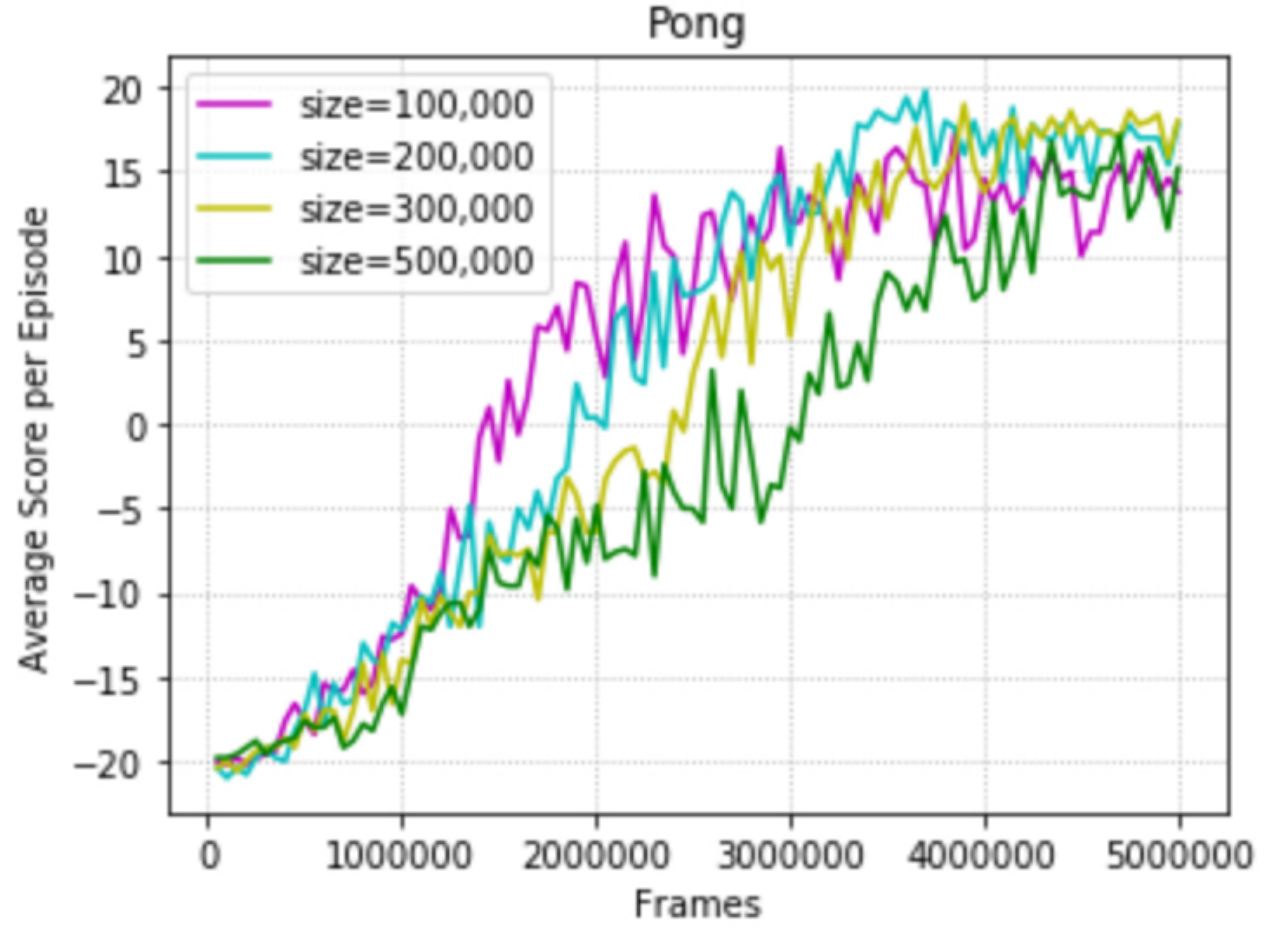}
  \caption{The first 5M steps in Figure \ref{Buffer}.}
  \label{BufferD} 
  \end{minipage}
\end{figure*}

\section{Implementation}

We use Pong of Atari 2600 provided with OpenAI Gym \cite{OpenAI} which can easily share the result of library and reinforcement learning algorithms.

Pong is the game of 21 points win. If the opponent cannot hit the ball, we get a reward of +1, and if we cannot, we get a reward of -1. That is, the reward set $R = \{-1, 0, 1 \} $.
Many of the algorithms targeting versatility such as DQN use Reward Clipping that clips rewards gained from games to $ [-1, 1]$.
It is a technique that we enables an agent to learn without changing other parameters.
However, Reward Clipping may make it impossible to distinguish between a high reward and a small reward with a large absolute value, so there is a possibility that it will not try to obtain high rewards\cite{Pop-art}.
We do not consider it in this research because the Pong's reward is $ r \in R $.


Like the DQN, we process one frame of the game such as Figure \ref{Pong} to $ 84 \times 84 $ and convert to grayscale and normalize it.
Then the state is set with the consecutive 4 frames together.
We show the main parameters and the network parameters setting for the experiment in Table \ref{tab1} and Table \ref{tab2} as Appendix.

We evaluate the learning speed of NEC2DQN, N-step DQN and Double DQN.
Also we observe the learning result by the difference in the size of Replay Buffer.
We test 5 times every 50,000 learning steps, and we use greedy policy $(\varepsilon = 0)$ in every test.

\section{Result}




Figure \ref{Others} is a comparison with NEC2DQN, N-step DQN $(N=10) $ and Double DQN.
Although Double DQN required more than 30M steps to earn more than 10 points, NEC2DQN achieved it in about 3M steps.
Furthermore, NEC2DQN has better learning efficiency and performance than N-step DQN.

Figure \ref{Buffer} and Figure \ref{BufferD} show the difference in results depending on Replay Buffer size.
The size affects the stability of sampling.
The larger it is, the better the performance is.
However, since the target value is also stored in Replay Buffer, the target value is too old to proceed well if it is too large like $ size = 500,000 $.
Looking at Figure \ref{BufferD}, the smaller size is, the faster learning is due to the newness of the target value at the 2M frame.
But as learning progresses, it turns out that the game score is not obtained well in the case the size is small like $ size = 100,000 $.
NEC2DQN needs considering this balance.

\section{Conclusion}

In this research, we have verified DQN is possible to use the good sampling efficiency of NEC in the beginning of learning using a Pong example.
We have showed that the learning speed is faster than only DQN by using the same and better target values for NEC and DQN.
Indeed, we observed a significant learning speed improvement by using NEC during 2M steps only.

However it is necessary to confirm whether this method succeeds also in other games.
Particularly, although NEC does not need to Reward Clipping, DQN is better to do Reward Clipping.
Thus, there is a possibility that we cannot use the NEC's advantage. 

Moreover, Since NEC and DQN do not connect as the same network, updating weights of the networks do not directly affect each other.
Therefore, it is easy to replace it with a network other deep Q-learning algorithm.
It is necessary to verify whether other deep Q-learning algorithms also works as well as DQN by changing $ Q_{DQN} $ used for $ Q_{N2D} $ to the Q-value of others.

The most important issue is that multiple networks learn while choosing appropriate target values.
In this paper, we have used NEC since it is excellent in Q-value estimation at the beginning of learning, but NEC is not necessarily useful for estimating Q-value.
Deep Q-learning from Demonstrations (DQfD)\cite{DQfD} is used to estimate the Q-value with reference to human play.
It is sometimes hard to go well in tasks that humans cannot do very well (such as Pong), but if it is a task that humans can do well, it is good for learning faster than NEC.
However, it is difficult to use human demos when we make our agent learn more complicated and time-consuming tasks.
It is important that various networks of deep Q-learning cooperate by learning while choosing an appropriate target value automatically.

\appendix

\begin{table}[htbp]
\caption{Hyper Parameter}
\begin{center}
\begin{tabular}{l | c}
\hline
\textbf{Parameter}&\textbf{Value}\\
\hline
Optimizer & RMSProp$(\varepsilon=0.01)^\dagger$ \\
Optimizer learning rate & 0.000025\\
Optimizer momentum & 0.95$^\dagger$ \\
Explore $\varepsilon$ & 1 $\to$ 0.01 over 1M steps\\
Replay buffer size & 300,000 \\
DND learning rate & 0.1\\
DND size & 500,000 per action$^{\dagger\dagger}$ \\
p for KDTree & 50$^{\dagger\dagger}$ \\
N-step returns $N$ & 10 \\
NEC embedding size & 64 \\
DQN embedding size & 512$^\dagger$ \\
Replay period & every 4 learning steps$^\dagger$ \\
Minibatch size & 32$^\dagger$ \\
Discount rate & 0.99$^\dagger$ \\
NEC2DQN change step $CS$ &  2M steps \\
\hline
\end{tabular}

$\dagger$same as Double DQN\cite{DDQN}\ \
$\dagger\dagger$same as NEC\cite{NEC}
\label{tab1}
\end{center}
\end{table}

\begin{table}[htbp]
\caption{Network}
\begin{center}
\begin{tabular}{l | c}
\hline
\textbf{Parameter}&\textbf{Value}\\
\hline
NEC,DQN: CNN channels & 32, 64, 64\\
NEC,DQN: CNN filter size & $8 \times 8, 4 \times 4, 3 \times 3$\\
NEC,DQN: CNN stride & 4, 2, 1\\
NEC : embedding size & 64\\
DQN : hidden layer & 512\\
DQN : output units & Number of actions\\
\hline
\end{tabular}
\label{tab2}
\end{center}
\end{table}


\begin{thebibliography}{00}
\bibitem{DQN} Mnih, Volodymyr, Kavukcuoglu, Koray, Silver, David, Rusu, Andrei A, Veness, Joel, Bellemare, Marc G, Graves, Alex, Riedmiller, Martin, Fidjeland, Andreas K, Ostrovski, Georg, et al. Human-level control through deep reinforcement learning. In Nature, 518(7540):529-533, 2015.
\bibitem{survey} Kai Arulkumaran, Marc Peter Deisenroth, Miles Brundage, and Anil Anthony Bharath. A Brief Survey of Deep Reinforcement Learning. arXiv:1708.05866, 2017.
\bibitem{ER} Long-Ji Lin. Self-Improving Reactive Agents Based on Reinforcement Learning, Planning and Teaching. Machine Learning, 8(3-4):293-321,1992.
\bibitem{PriER}Schaul, Tom Schaul, John Quan, Ioannis Antonoglou, and David Silver. Prioritized Experience Replay. In ICLR, 2016.
\bibitem{NEC} Alexander Pritzel, Benigno Uria, Sriram Srinivasan, Adria` Puigdome`nech, Oriol Vinyals, Demis Hassabis, Daan Wierstra, and Charles Blundell. Neural Episodic Control. arXiv:1703.01988, 2017.
\bibitem{bellman} Richard Bellman. On the Theory of Dynamic Programming. In PNAS, 38(8):716-719, 1952.
\bibitem{Q-learn} Christopher JCH Watkins and Peter Dayan. Q-Learning. Machine Learning, 8(3-4):279-292, 1992.
\bibitem{CNN} Alex Krizhevsky, Ilya Sutskever, Geoffrey E,Hinton. ImageNet classification with deep convolutional neural networks. In ANIPS, 2012.
\bibitem{Double Q-learn} Hado van Hasselt. Double Q-Learning. In NIPS, 2010.
\bibitem{DDQN} Hado van Hasselt, Arthur Guez, and David Silver. Deep Reinforcement Learning with Double Q-Learning. In AAAI, 2016.
\bibitem{MFEC} Charles Blundell, Benigno Uria, Alexander Pritzel, Yazhe Li, Avraham Ruderman, Joel Z Leibo, Jack Rae, Daan Wierstra, and Demis Hassabis. Model-Free Episodic Control. arXiv:1606.04460, 2016.
\bibitem{kd} Bentley, Jon Louis. Multidimensional binary search trees used for associative searching. Commun. ACM, 18(9): 509-517, 1975.
\bibitem{multi-step} Jing Peng and Ronald Williams. Incremental multi-step Q-learning. Machine Learning, 22:283-
290, 1996.
\bibitem{N-step Safe} Remi Munos, Tom Stepleton, Anna Harutyunyan, and Marc G Bellemare. Safe and Efficient Off-Policy Reinforcement Learning. In NIPS, 2016.
\bibitem{Keio} Shoya Matsumori, Takuma Seno, Toshiki Kikuchi, Yusuke Takimoto, Masahiko Osawa, Michita Imai. Embedding Cognitive Map in Neural Episodic Control. In SIG-AGI, 2017
\bibitem{OpenAI} Greg Brockman, Vicki Cheung, Ludwig Pettersson, Jonas Schneider, John Schulman, Jie Tang, and Wojciech Zaremba. OpenAI Gym. arXiv:1606.01540, 2016.
\bibitem{Pop-art} Hado van Hasselt, Arthur Guez, Matteo Hessel, Volodymyr Mnih, David Silver. Learning values across many orders of magnitudes. In NIPS, 2016.
\bibitem{DQfD} Todd Hester, Matej Vecerik, Olivier Pietquin, Marc Lanctot, Tom Schaul, Bilal Piot, Andrew Sendonaris, Gabriel Dulac-Arnold, Ian Osband, John Agapiou, et al. Deep Q-learning from Demonstrations.  arXiv:1704.03732, 2017.
\end{thebibliography}
\end{document}